\documentclass[9pt,twocolumn,twoside]{pnas-new}

\templatetype{pnasresearcharticle} 

\usepackage{overpic}
\renewcommand{\d}{\mathrm{d}}
\newcommand{\norm}[1]{\left\lVert #1\right\rVert}

\title{Extended critical regimes of deep neural networks}

\author[a,1]{Cheng Kevin Qu}
\author[a,1]{Asem Wardak} 
\author[a]{Pulin Gong}

\affil[a]{School of Physics\\
          University of Sydney\\
          NSW 2006\\
          Australia
          }

\leadauthor{Lead author last name} 

\authorcontributions{Author contributions: C.K., A.W. and P.G. designed research, performed research and wrote the paper.}
\authordeclaration{The authors declare no competing interest.}
\equalauthors{\textsuperscript{1}C.Q. and A.W. (Author Two) contributed equally to this work.}
\correspondingauthor{\textsuperscript{2}To whom correspondence should be addressed. E-mail: pulin.gong@sydney.edu.au}


\begin{abstract}

Deep neural networks (DNNs) have been successfully applied to many real-world problems, but a complete understanding of their dynamical and computational principles is still lacking.
Conventional theoretical frameworks for analysing DNNs often assume random networks with coupling weights obeying Gaussian statistics.
However, non-Gaussian, heavy-tailed coupling is a ubiquitous phenomenon in DNNs.
Here, by weaving together theories of heavy-tailed random matrices and non-equilibrium statistical physics, we develop a new type of mean field theory for DNNs which predicts that heavy-tailed weights enable the emergence of an extended critical regime without fine-tuning parameters. 
In this extended critical regime, DNNs exhibit rich and complex propagation dynamics across layers.
We further elucidate that the extended criticality endows DNNs with profound computational advantages: balancing the contraction as well as expansion of internal neural representations and speeding up training processes, hence providing a theoretical guide for the design of efficient neural architectures.    

\end{abstract}

\dates{This manuscript was compiled on \today}
\doi{\url{www.pnas.org/cgi/doi/10.1073/pnas.XXXXXXXXXX}}

\begin{document}

\maketitle
\thispagestyle{firststyle}
\ifthenelse{\boolean{shortarticle}}{\ifthenelse{\boolean{singlecolumn}}{\abscontentformatted}{\abscontent}}{}

\dropcap{D}eep neural networks (DNNs) have achieved remarkable success over the past decade across a variety of architectures in fields including visual object classification \cite{Krizhevsky2012}, natural language processing \cite{NEURIPS2020_1457c0d6} and speech recognition \cite{yu2014automatic}.
In these systems, input data passes through large numbers of hidden layers composed of neurons, giving rise to complex activity dynamics in order to produce highly abstract representations of concepts needed for real-world classification tasks.
Understanding how such complex dynamics allow deep neural networks to learn and compute is thus a longstanding topic of interest in artificial intelligence and has implications for other similar complex systems such as the brain \cite{Chialvo2010}.
Since the deep learning revolution of the past decade \cite{LeCun2015}, a classical theoretical framework has emerged using a variety of physical and mathematical tools including statistical physics and random matrix theory in order to understand the dynamics of signal propagation through deep networks \cite{Bahri2020}.
These works have made use of Gaussian mean-field approaches 
to establish parameter phases of vanishing and exploding signal propagation \cite{Poole2016,Pennington2018}, corresponding to the classical phases of order and chaos respectively \cite{Schoenholz2016}.
These important insights have informed initialisation strategies which lead to the faster and more successful training of deep neural networks, such as concentrating the eigenvalues of the initial weight matrix around the unit circle \cite{Glorot2010}.

A majority of theoretical studies have generally assumed Gaussian random neural networks or Gaussian initialisations.
However, it has been empirically demonstrated that deep networks in practice have heterogeneous, heavy-tailed coupling regardless of architecture \cite{simsekli19a, HodgkinsonM21}.
Moreover, the generalisation abilities of deep networks are closely linked with the heavy-tailed properties of the singular values of its weight matrices \cite{Martin2021}.
These empirical observations raise fundamental questions of whether and how heavy-tailed heterogeneity endows DNNs with crucial dynamical and computational properties.
Answering these questions would provide a better understanding of the functional mechanisms of DNNs, and help guide the future design of such systems.

Here, by linking non-equilibrium statistical physics with recent results in heavy-tailed random matrix theory, we develop a novel mean-field theory for random deep neural networks with heavy-tailed connectivity.
This theory reveals an extended critical phase where DNNs balance order and disorder in a broad region of parameter space, allowing for deep information propagation; fine-tuning of parameters is thus no longer required as predicted by conventional theories.
We analytically demonstrate the presence of this extended critical phase by examining the heavy-tailed layerwise Jacobian matrices of DNNs, and validate our analytical results numerically. 
Based on our theory, we formulate a phase diagram which can be used as a guide for designing DNNs in practice.
To demonstrate this, we select initialisation parameters within the extended critical regime and find the network displays faster training as well as superior generalisation on unseen data.
In addition, we reveal that the computational advantages of balancing compression and expansion emerge generically in the extended critical regime. 
These results indicate that our theory provides a framework for understanding dynamical and computational properties of DNNs.


\section{Results}

We first demonstrate that empirical coupling weights of DNNs have a heavy-tailed distribution which is ubiquitous across network architectures.
We then formulate a new dynamical mean-field theory revealing a novel phase diagram of DNNs, which features a broad, extended critical regime adjacent to the classical silent and chaotic regimes.
We numerically validate the theoretically predicted properties of the extended regime and elucidate its computational advantages by demonstrating that extended criticality enables the balance of compression and expansion of internal representations corresponding to neural inputs.

\subsection{Heavy-tailed coupling weights of pretrained neural networks}
\label{sec:empirical}

\begin{figure*}[ht]
\centering
\includegraphics[width=17.8cm]{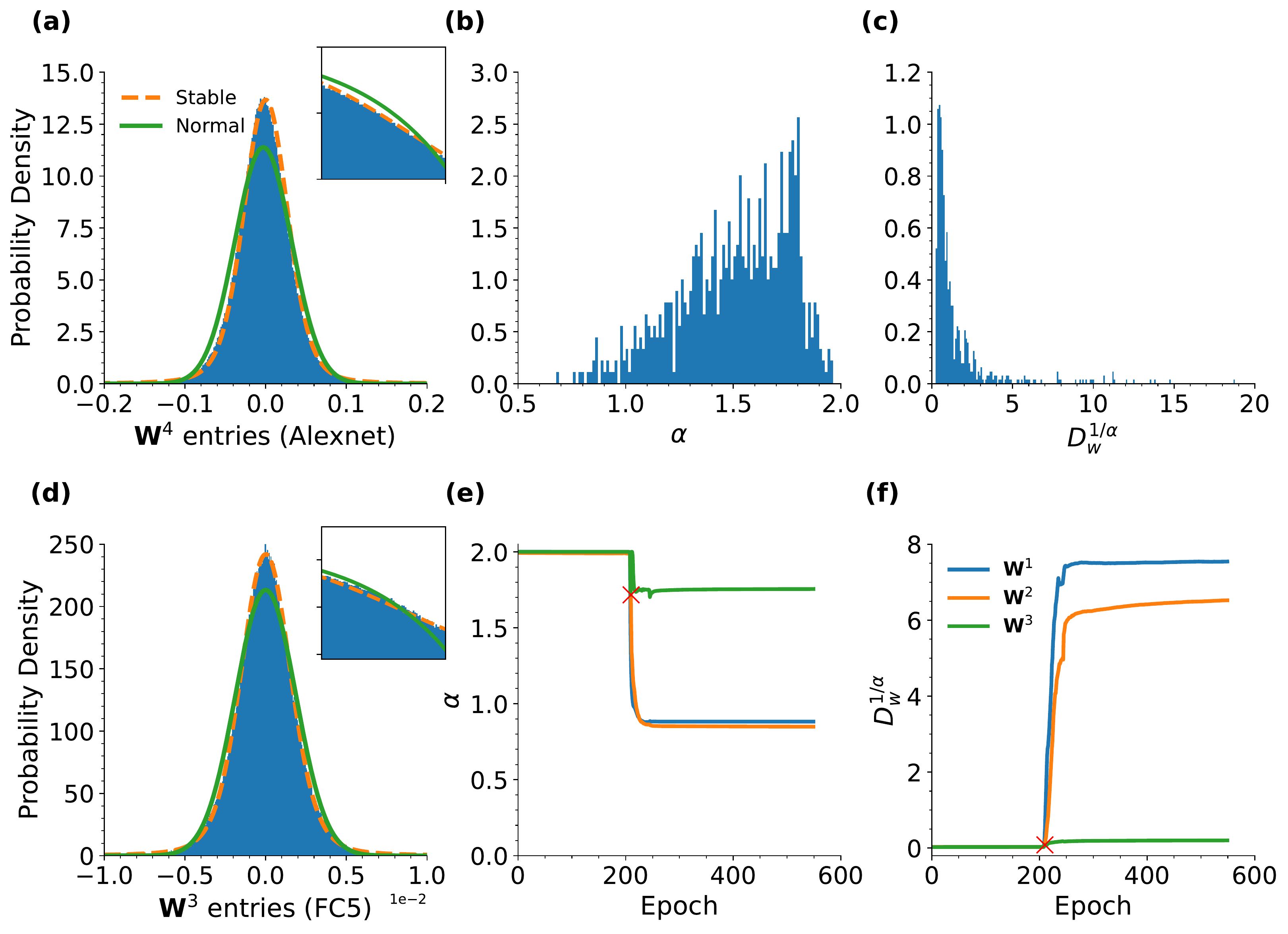}
\caption{\label{fig:pretrained_torch_plaw} \textbf{Stable fit of weight matrix entries.}
(a) Entries of a selected $\mathbf{W}^4$ from a pretrained \emph{AlexNet} with over $6.5 \times 10^5$ entries fitted to stable and Gaussian distributions via maximum likelihood (log-log tail in inset).
(b) The distribution of the stability parameter corresponding to each of 699 weight matrices from 10 pretrained networks.
(c) Same as in (b) but for the normalised scale parameter $D_w^{1/\alpha}$.
(d) Same as in (a) but for entries of a selected $\mathbf{W}^3$ from FC5 with over $6 \times 10^5$ entries at epoch 210.
(e) The evolution of the stability parameter throughout 550 epochs of training for FC5 with Gaussian initialisation. The red cross represents the epoch featured in (d), with the fitted parameters estimated at the end of each epoch.
(f) Same as in (e) but for the scale parameter.
}
\end{figure*}


It has recently been shown that across a wide variety of architectures, the spectrum of weight matrices of pretrained DNNs are heavy-tailed (i.e. they have power-law tails) \cite{Martin2021}. Such heavy-tailed statistics could arise from correlations within the weight matrices, or the weight entries themselves originating from a heavy-tailed distribution.
We investigate the latter hypothesis using pretrained network weight matrices from the Pytorch library (version 1.6.0).
Specifically, we fit the entries of each weight matrix $\mathbf{W}^l$ (excluding the bias and batch-normalization layers as in \cite{Martin2021}) between the $(l - 1)^{\text{th}}$ and $l^{\text{th}}$ layers individually as a L\'{e}vy $\alpha$-stable distribution (often termed \emph{stable distribution}) \cite{Nolan2020}, which is defined by a characteristic function involving a tuple $(\alpha, \beta, \sigma, \mu)$ containing the stable, skewness, scale and location parameters respectively,
\begin{equation} \label{eq: stable_dist}
\varphi(u;\alpha,\beta,\sigma,\mu) = \exp\left( - \vert \sigma u \vert^{\alpha}(1 - i \beta \text{sgn}(u)\Phi(u; \alpha)) + iu\mu \right)
\end{equation}
where $\text{sgn}(u)$ is the sign of $u$ and
$$
\begin{aligned}
\Phi(u; \alpha) = 
\begin{cases}
\tan\left( \frac{\pi \alpha}{2} \right) \quad &\alpha \neq 1 \\
-\frac{2}{\pi}\log \vert u \vert &\alpha = 1
\end{cases}.
\end{aligned}
$$
If a random variable $X$ is drawn from a stable distribution, we denote it by $X \sim S_{\alpha}(\beta,\sigma,\mu)$, and $X \sim S_{\alpha}(\sigma)$ if it is symmetric with $\beta = \mu = 0$ \cite{Samorodnitsky_Taqqu}.
Apart from Gaussian distributions for which $\alpha=2$, all stable distributions with $\alpha<2$ have power-law tails, which is why $\alpha$ is also referred to as the tail index.
As an example, we present the distribution of entries of a weight matrix from a pretrained convolutional architecture (\emph{AlexNet}) in Fig.\ \ref{fig:pretrained_torch_plaw}(a), which can be better fitted to an $\alpha$-stable distribution with $\alpha<2$ compared to a Gaussian distribution.
We perform the same analysis to 699 weight matrices across 10 pretrained networks from Pytorch and plot the distribution of tail indices $\alpha$ (Fig.~\ref{fig:pretrained_torch_plaw}(b)); over 99.72\% of the tail indices are lower than 1.95.
In Fig.\ \ref{fig:pretrained_torch_plaw}(c), the corresponding normalised scale parameters $D_w=2\sqrt{N_w N_h}\sigma^\alpha$ are consistent with the range of the scale parameter ($D_w^{1/\alpha}$) in the extended critical regime described below (Fig.~\ref{fig:phase_transition}).\footnote{Here $N_w$ and $N_h$ denote the width and height of the corresponding weight matrix.}
Shaprio-Wilk tests reveal that 99.57\% of the weight matrices reject the null hypothesis of Gaussian weights at the significance level of 2.5\%.
Additionally, all of the p-value ratios of the Kolmogorov–Smirnov (KS) test with respect to the maximum likelihood fit to the stable and normal distribution are greater than or equal to 1.
Although a perfect fit to the stable distribution is hindered by finite-size effects, it is not necessary for our L\'evy mean-field analysis because the theory only requires the assumption of a power-law tail in the large network limit.

Similarly, we demonstrate that such heavy-tailed weights can emerge during the training process.
We train a fully-connected feedforward neural network with Gaussian initialisations. 
In particular, this network which consists of 4 hidden layers (\emph{FC5}) is optimised via standard SGD for 550 epochs with learning rate 0.1 and batch size 128.
We again provide a direct comparison between a Gaussian and stable fit for the entries of $\mathbf{W}^3$ in Fig.~\ref{fig:pretrained_torch_plaw}(d).
The stability and scale parameters of the first 3 weight matrices were tracked during the process (Fig.\ \ref{fig:pretrained_torch_plaw}(e,f)).
After 550 epochs, the network attains a training (testing) accuracy of 99.92\% (98.21\%) on the MNIST dataset.
In the course of training, the weights remain Gaussian for approximately 200 epochs, and then deviate to a stable distribution with $\alpha < 2$ where similar changes of the scale parameter $\sigma$ take place simultaneously.
Towards the end of epoch 220 the distributions of $\mathbf{W}^1$ to $\mathbf{W}^3$ stabilise to heavy-tailed distributions (Fig.\ \ref{fig:pretrained_torch_plaw}(e)).




\subsection{L\'evy mean-field theory}

Consider a feedforward neural network of depth $L$ in the wide limit, so that for simplicity each layer $l$ has $N$ neurons described by a neural activity vector $\mathbf{x}^l$ along with an $N\times N$ weight matrix $\mathbf{W}^l$.
The propagation dynamics arising from the input $\mathbf{x}^0\in\mathbb{R}^N$ is given by
\begin{equation}\label{eq: network_fc}
    \mathbf{x}^l = \phi(\mathbf{h}^l) ~,\qquad
    \mathbf{h}^l = \mathbf{W}^l \mathbf{x}^{l-1} + \mathbf{b}^l ~,
\end{equation}
where $\mathbf{h}^l$ is the input at layer $l$, $\mathbf{b}^l$ is a bias vector, and $\phi$ is an element-wise nonlinear activation function.
Connecting our results in Section \ref{sec:empirical} with the finding that the singular spectrum of weight matrices from pretrained DNNs are heavy-tailed \cite{Martin2021}, leads to a standard result in random matrix theory asserting that the singular values of matrices with independent, heavy-tailed entries are themselves heavy-tailed \cite{Bordenave2011}.
Hence, to understand the impact of heavy-tailed weights on network dynamics, we stipulate that the weights and biases be independent and identically distributed (i.i.d.) as
$\mathbf{W}_{ij}^l\sim S_{\alpha}((D_w/2N)^{1/\alpha})$ and
$\mathbf{b}_i^l\sim S_{\alpha}((D_b/2)^{1/\alpha})$ 
where $1\leq\alpha\leq2$, with $D_w$ and $D_b$ parametrising the scale of the weight and bias respectively.
The classical setup is recovered in the $\alpha\to2$ limit with $D_{\{w,b\}}=\sigma_{\{w,b\}}^2$ \cite{Poole2016}.
The following derivations and results also hold when the independent weights are distributed with the same asymptotic heavy tail as a stable distribution, namely
\begin{equation}
p_{\mathbf{W}_{ij}^l}(x)
\sim \frac{c_\alpha D_w}{2N} |x|^{-1 - \alpha}
\end{equation}
for the probability density function $p(x)$, where $c_\alpha:=\Gamma(1+\alpha)\sin(\pi\alpha/2)/\pi$.
Applying the generalised central limit theorem \cite{Samorodnitsky_Taqqu, Wardak2021} to Eq.\ (\ref{eq: network_fc}) in the limit $N\to\infty$ yields the convergence of the inputs $\mathbf{h}_i^l$ over neurons $i$ to a stable random variable in distribution,
\begin{equation}\label{eq:GCLT}
\begin{aligned}
    \mathbf{h}_i^l &\stackrel{i}{\sim} S_{\alpha}\left[ \left(\frac{D_w}{2N} \sum_{j=1}^{N}\left|\phi\left(\mathbf{h}_{j}^{l-1}\right)\right|^{\alpha}
    +
    \frac{D_b}{2}\right)^{1 / \alpha} \right]
    \\
    &= S_{\alpha}\left[ \left(\frac{D_w}{2}
    \int |\phi(z)|^\alpha p_{\mathbf{h}^{l-1}}(z) \d z
    +
    \frac{D_b}{2}\right)^{1 / \alpha} \right] ~.
\end{aligned}
\end{equation}
Parameterising the distribution of the neural input $\mathbf{h}^l_i\sim S_\alpha((q^l/2)^{1/\alpha})$ at layer $l$ by $q^l$, we obtain a L\'evy mean-field iterative map for $q^l$ from Eq.\ (\ref{eq:GCLT}) by repeatedly applying the above derivation of stable distributions,
\begin{equation} \label{eq: iter_map}
    q^l 
    = D_w \int |\phi(z)|^{\alpha} p_{S_{\alpha}\left((q^{l-1}/2)^{1/\alpha} \right)}(z) \d z + D_b
\end{equation}
for $l=2,\dots,L$, where the initial condition is $q^1 = D_w q^0 + D_b$ and $q^0=\sum_i |\mathbf{x}^0_i|^\alpha / N$ is the $\alpha$-th moment of the initial activity layer.
The parameter $q^l$ characterises the fluctuations of the neural input distribution at layer $l$ and reduces to the classical normalised squared length when $\alpha=2$ \cite{Poole2016}.

Eq.\ (\ref{eq: iter_map}) constitutes the L\'evy mean-field equation for the feedforward neural network.
The fixed point is then obtained by setting $q^{l-1}=q^l=q^*$ for large $l$; in order to guarantee the finiteness of the right-hand side of the equation, we assume that $\phi(|x|) = o(|x|)$ using the little-o notation, which includes all sigmoidal functions.
Linearising the network around the fixed point $q^*$ yields a random input-output Jacobian,
\begin{equation} \label{eq:layer_jac}
    \frac{\partial\mathbf{x}^L}{\partial\mathbf{x}^0}
    = \prod_{l=1}^L \mathbf{D}^l\mathbf{W}^l
\end{equation}
where $\mathbf{D}^l$ is a diagonal matrix with entries $\mathbf{D}^l_{ij}=\phi'(\mathbf{h}_i^l)\delta_{ij}$.
Leveraging the statistical properties of the eigenvectors and eigenvalues of the Jacobian as well as its constituent layerwise Jacobians $\mathbf{D}^l\mathbf{W}^l$ gives us a method to consistently characterise the onset of edge-of-chaos criticality.


\subsection{Jacobian operator of heavy-tailed deep neural networks}

The conventional Gaussian mean-field approach characterises the transition to chaos by computing the covariance of two inputs as they propagate through the layers of DNNs \cite{Schoenholz2016,Poole2016,lee2018deep,Pennington2018}.
In these studies, chaos is characterised by the separation of nearby points as they propagate through the layers, with asymptotic expansions yielding depth scales over which information may approximately propagate as the magnitude of a single input or the correlation between two inputs.
However, because the variances of such inputs become infinite upon propagation by a single layer for heavy-tailed networks, these covariances are no longer guaranteed to be well-defined in heavy-tailed deep networks.
To circumvent this, we examine the statistical propagation of eigenvectors of layerwise Jacobian matrices of the form $\mathbf{D}^l \mathbf{W}^l$ and $\mathbf{W}^{l+1} \mathbf{D}^{l}$.\footnote{These two matrices correspond to the post- and pre- activations layerwise Jacobians.}
The Jacobian operators of Gaussian random neural networks satisfy a circular law with a finite spectral radius \cite{Rajan2006, Aljadeff2015} and delocalised eigenvectors \cite{Evers2008} that spread evenly across the neural sites.
Traditionally, examination of the spectral radius provides identification of transition to chaos, i.e. if the maximum singular value of the Jacobian crosses unity, signal propagation expands space and vice versa; hence allowing us to differentiate between an ordered and chaotic regime.
Because the maximum singular value of heavy-tailed matrices is infinite \cite{Bordenave2011},
we apply a recent theory by our group \cite{Wardak2022} which is particularly powerful for analysing the layerwise Jacobian matrices of heavy-tailed DNNs.

Ref.\ \cite{Wardak2022} demonstrates the key properties of a time-varying Jacobian operator around the stationary state in a recurrent neural network (RNN) context, which is equal to the layerwise Jacobian with form $\mathbf{W}\mathbf{D}$ of our feedforward network around the fixed point due to Eq.\ (\ref{eq: network_fc}).
Exploiting the locally treelike properties of heavy-tailed random matrices, a cavity approach is applied \cite{Bordenave2012, LucasMetz2019, Wardak2022} to find that its eigenvalue density $\rho(z)$ has infinite support with an exponential cutoff at large modulus \cite{Bordenave2011}, such that
\begin{align}\label{eq:RMT_eigenvalues}
    \rho(z) = \frac{y_*^2 - 2|z|^2 y_* \partial_{|z|^2}y_*}{\pi} \left\langle \frac{|\chi_i|^2 SS'}{(|z|^2 + |\chi_i|^2 y_*^2 SS')^2} \right\rangle_i
\end{align}
where $\langle..\rangle_i$ denotes averaging over $i$ and any relevant random variables, $\chi_i=D_w^{1/\alpha}\phi'(h_i)$ varies over neurons, $S,S'\sim S(\alpha/2,1,0,(c_\alpha/4c_{\alpha/2})^{2/\alpha})$ are independent, skewed stable random samples, and $y_*$ is found by solving the equation
\begin{align}
    1 = \left\langle\left(\frac{|\chi_i|^2 S}{|z|^2 + y_*^2 |\chi_i|^2 S S'}\right)^{\alpha/2}\right\rangle_i ~.
\end{align}
Conventional approaches for the edge-of-chaos transition would thus conclude that heavy-tailed networks are always chaotic, ignoring the effect of the exponential cutoff in practice.

Using the cavity approach \cite{Wardak2022} shows that the right eigenvectors of the layerwise Jacobian for $\phi=\tanh$ are spatially multifractal over neural sites with a mixture of localised and delocalised properties.
This is proven by deriving the localisation of the left and right eigenvectors in terms of the inverse participation ratio $\mathrm{IPR}_q(v) = \sum_i |v_i|^{2q}$.
Such multifractal localisation over neurons is defined by a nontrivial dependence on $q$ of the (generalised) fractal dimension $D_q$ appearing in the inverse participation ratio \cite{Evers2008}
\begin{equation} \label{eq:frac_dim}
    \mathrm{IPR}_q(v)\sim N^{(1-q)D_q}
\end{equation} 
for large system size $N$, where $D_q=0$ ($1$) corresponds to localised (delocalised) spatial profiles over neurons.
Based on Eq.\ (\ref{eq:frac_dim}), the fractal dimension corresponding to the (eigen)vector can be estimated via the asymptotic relation $D_q \sim \log_N \text{IPR}_q(v)/(1 - q)$.
The localisation and delocalisation properties inherent to the multifractal behaviour may respectively enable dynamical balancing of dimensional compression and expansion of robust internal neural representations as observed in trained RNNs \cite{Farrell2019}.

\subsection{An extended critical regime of signal propagation in deep neural networks}

We next demonstrate that the independence of Jacobian eigenstate statistics from the phase of the complex eigenvalue in Eq.\ (\ref{eq:RMT_eigenvalues}) allows us to develop a rigorous characterisation of criticality and the transition to chaos, which remains consistent with the Gaussian case.

\begin{figure*}
\centering
\begin{tabular}[b]{cc}
\begin{tabular}[b]{c}
\begin{overpic}[width=.3\textwidth]{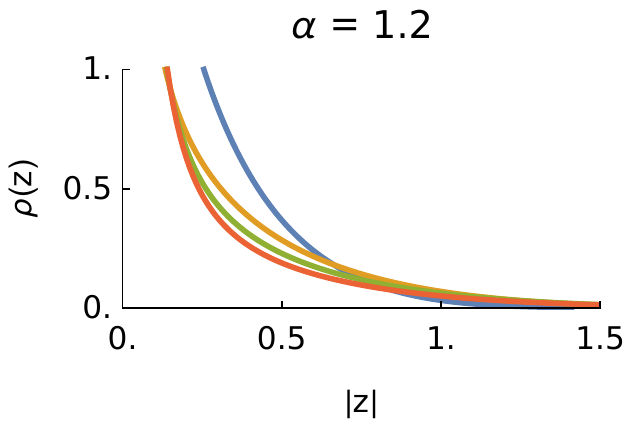}
\put (0,70) {\large\textbf{(a)}}
\put (60,20) {\includegraphics[width=.15\textwidth]{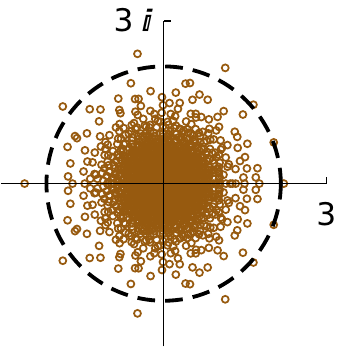}}
\put (60,-50) {\includegraphics[width=.15\textwidth]{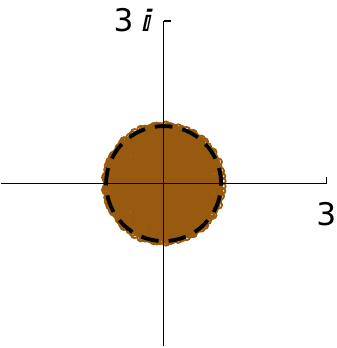}}
\end{overpic}\\
\includegraphics[width=.3\textwidth]{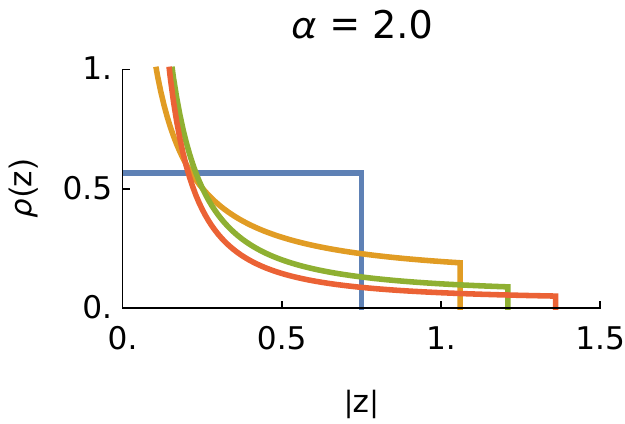}
\end{tabular} &
\begin{overpic}[width=.49\textwidth]{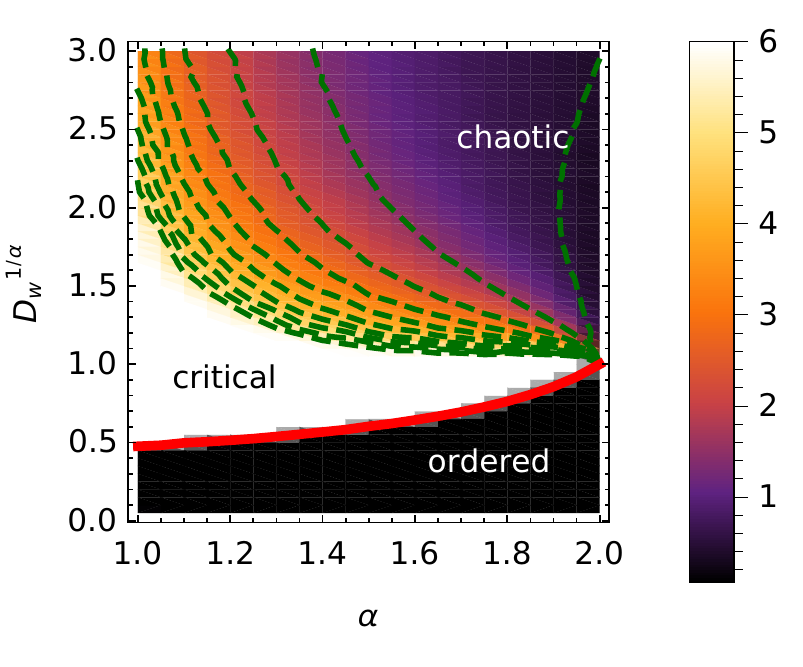}
\put (0,85) {\large\textbf{(b)}}
\end{overpic}
\end{tabular}
\caption{\label{fig:phase_transition}
\textbf{Jacobian random matrix theory and phase diagram.}
(a)
The density of states of the layerwise Jacobian $\mathbf{W}\mathbf{D}$ for deep random networks with zero bias and $D_w^{1/\alpha}=0.75,\dots,3$ (blue, yellow, green, red).
Insets depict the empirical eigenvalue distribution and characteristic spectral radius beyond which exponential suppression of eigenvalue density dominates for $D_w^{1/\alpha}=1.5$, equal to the true spectral radius when $\alpha=2$.
(b)
The $\alpha$---$D_w^{1/\alpha}$ phase transition diagram for heavy-tailed deep random networks with $\phi=\tanh$.
The colors depict the maximum $L$ such that the ratio of Jacobian averages $\mathcal{J}_{f_1}$ using $f_1(r)=\mathrm{sgn}(\log(r))|\log(r)|^L$ between the given point and the edge-of-chaos ordered transition (red line) at that $\alpha$ remains greater than unity, with a cutoff at $6$.
The green lines represent the corresponding points at which this ratio equals unity using $f_2(r)=(r-2)^L$ and $L=1,11,21,\dots,101$ (starting from the right), with the enclosed region performing better than the edge-of-chaos ordered transition line with respect to the Jacobian average $\mathcal{J}_{f_2}$.
}
\end{figure*}

Given that all eigenvectors of the layerwise Jacobian with a fixed eigenvalue modulus have the same localisation characteristics, we may deduce by symmetry that large randomly selected matrices from the ensemble have as eigenvectors all normalised vectors with these localisation properties.
To illustrate this, observe that 
fixing the IPR for all $q$ also fixes the distribution of eigenvector entry magnitudes.
Neglecting eigenvector correlations, the full network Jacobian is thus expected to yield an eigenvalue with modulus $|\lambda|^L$ for the direction corresponding to $v$ with layerwise Jacobian eigenvalue $\lambda$.\footnote{
Such an argument only applies close to the fixed point where one can linearise the network dynamics;
as the network moves further from the stationary state, the localisation properties corresponding to a given eigenvector also change to the point that the direction is no longer necessarily an eigenvector, and nonlinear behaviour becomes significant.
}
The local propagation of signals in a random direction through the deep network is thus determined by the proportion of eigenvalues residing away from zero, which is unknowable solely from the spectral radius of the operator.
Remarkably, we find that deep in the classical chaotic regime, most eigenvalues reside closer to zero despite the spectral radius being larger; Fig.\ \ref{fig:phase_transition}(a) shows the eigenvalues of the layerwise Jacobian for heavy-tailed ($\alpha=1.2$) and Gaussian ($\alpha=2.0$) random DNNs around the fixed point.
The predominance of Jacobian eigenvalues close to zero results in the inability of information in a given direction to be faithfully propagated through the network in a manner which is distinguishable from noise.
On the other hand, 
a larger proportion of eigenvalues residing away from zero results in the maintenance of signal propagation in a particular eigendirection through more layers in the random deep network,
improving generalisability and resulting in correlated, edge-of-chaos behaviour.

To rigorously establish the link between signal propagation and the Jacobian eigenvalue density, we compute averages over Jacobian eigenvalues via
\begin{align}\label{eq:jacavg}
    \mathcal{J}_f := \langle f(|\lambda_i|) \rangle
    = \int_{\mathbb{C}} f(|z|) \rho(z) \, dz
\end{align}
where $f$ is an increasing function which penalises small eigenvalues and rewards eigenvalues with large modulus.
The local signal propagation ability of the deep network can then be expressed using its Jacobian average.
To compare the local signal propagation abilities between networks with different $\alpha$, we first compute the Jacobian average at the ordered transition line $(\alpha,\overline{D_w})$ at which the fixed point becomes non-negligible ($q^*\sim0.01$, Fig.\ \ref{fig:phase_transition}(b) red line).
The corresponding ordered phase corresponds with the region where Jacobian eigenvalues with modulus greater than unity are exponentially suppressed in probability.
We then compute the dimensionless ratio of Jacobian averages (Eq.\ (\ref{eq:jacavg})) at parameters $(\alpha,D_w)$ with their values at the ordered transition $(\alpha,\overline{D_w})$.

Evaluating the Jacobian average for Gaussian DNNs ($\alpha=2$) shows that it is maximal at the classical edge-of-chaos transition, $D_w=1$, due to the concentration of eigenvalues around zero in the chaotic regime despite a larger spectral radius.
Consequently our characterisation of deep information propagation is consistent with those reported elsewhere for Gaussian DNNs \cite{Schoenholz2016}.
More importantly, for $\alpha<2$ we find an extended region in phase space where the ratio of the Jacobian average with respect to the ordered transition is greater than 1 (Fig.\ \ref{fig:phase_transition}(b)), indicating signal propagation through more layers in the network relative to the edge-of-chaos ordered transition point (criticality).
To determine the continuous nature of the transition to chaos, we compute the size of this extended critical region using monotonic averaging functions $f$ which progressively become more discriminatory between small and large eigenvalue moduli with greater depth $L$, such as $f_1(r)=\mathrm{sgn}(\log(r))|\log(r)|^L$ (Fig.\ \ref{fig:phase_transition}(b), coloured) and $f_2(r) = (r-2)^L$ (Fig.\ \ref{fig:phase_transition}(b), green lines).
Employing greater values of $L$ in the Jacobian average serves as a proxy for information being able to penetrate greater numbers of layers in the deep network.
Importantly, our results remain robust to changes in the form of the Jacobian average as long as $f$ is increasing.
In heavy-tailed networks, the chaotic phase continuously transitions into a critical regime where the ratio of Jacobian averages remains greater than unity even for large $L$, predicting superior propagation of information through deep networks compared to the edge-of-chaos transition line.
This network phase of edge-of-chaos criticality exists in an extended region of nonzero area in parameter space $(\alpha, D_w)$.
Regardless of the specific Jacobian average used, the extended critical regime closes into the classical edge-of-chaos point $D_w=1$ in the Gaussian limit, $\alpha=2$.

The presence of multifractal eigenvectors in the layerwise Jacobians illustrated above distinguish the extended criticality of our theory from other schemes such as hierarchical modular networks with Griffiths phases \cite{Moretti2013}.
Moreover, Griffiths phases are clearly separated from the inactive and active phases by a first-order phase transition at the critical spreading rate; the extended critical regime instead exhibits a second-order continuous transition with the active chaotic phase such that the crossover region in phase space does not diminish with increasing network size.
A continuous transition also exists between the extended critical and ordered phases, parameterised by the cutoff of exponential suppression in the eigenvalue density.


\subsection{Preservation of multifractality during training}

\begin{figure}[ht]
\centering
\begin{overpic}[width=0.45\textwidth]{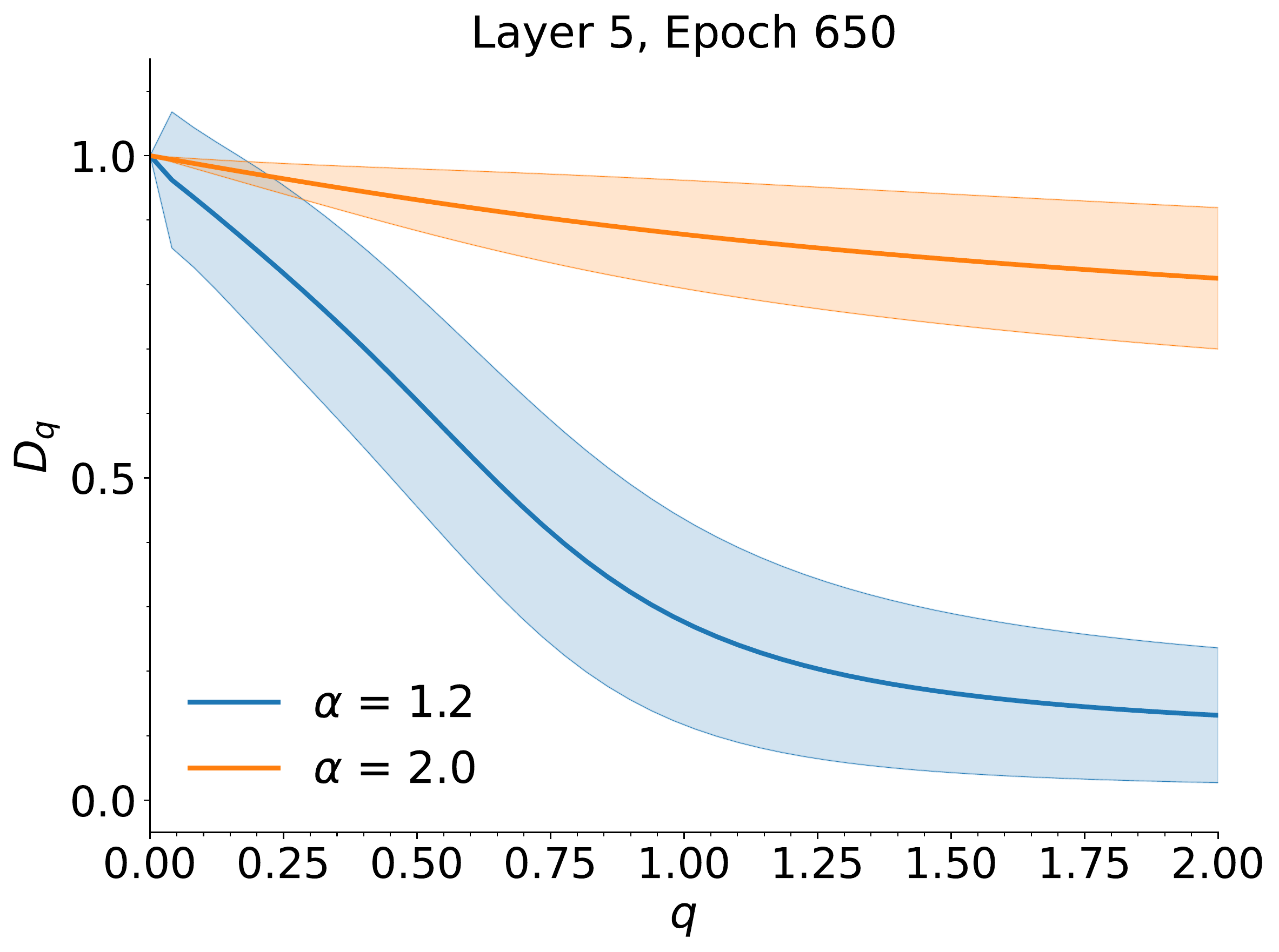}
\put (0,72) {\large\textbf{(a)}}
\end{overpic} \\
\begin{overpic}[width=0.45\textwidth]{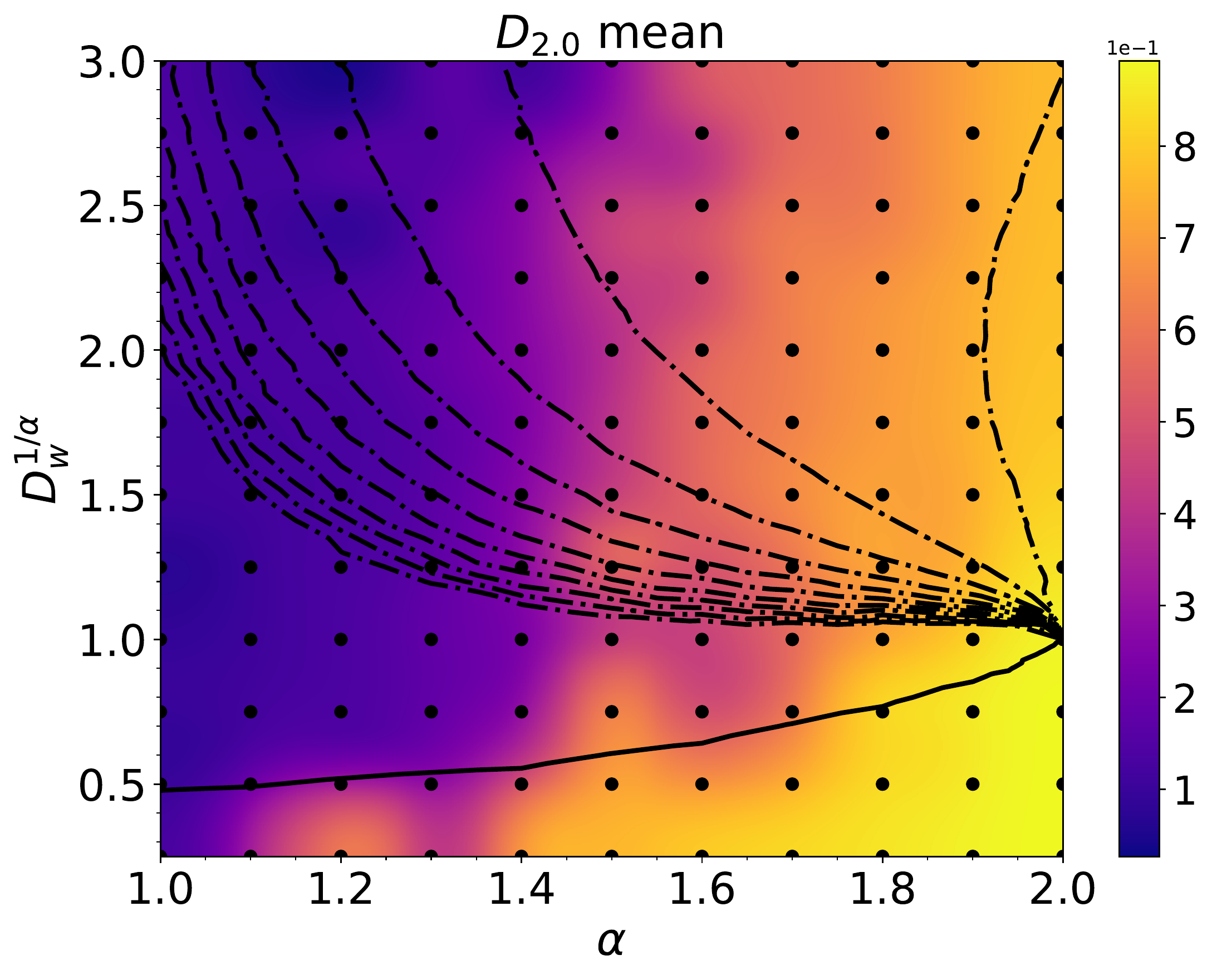}
\put (0,79) {\large\textbf{(b)}}
\end{overpic}
\caption{\label{fig:dq_jac_transition_l=4_epoch=650} \textbf{Fractal dimension $D_q$ vs $q$}. 
(a) The mean fractal dimension $D_q \sim \log_N \text{IPR}_q(v)/(1 - q)$ of the right eigenvectors of $\mathbf{W}^6 \mathbf{D}^5$ plotted with its standard deviation respectively for $\alpha = 1.2$ and $\alpha = 2.0$ with $D_w^{1/\alpha} = 1.5$ at epoch 650 in both cases.
(b) The mean fractal dimension across all right eigenvectors of $\mathbf{W}^6 \mathbf{D}^5$ at epoch 650 plotted on the phase overlaid with the phase transition diagram in Fig.~\ref{fig:phase_transition},
}
\end{figure}

As multifractality is more superior compared to delocalisation for separating crucial features from random noises shown in Section \ref{sec:balance}, it is of great significance to study if this characteristic is maintained during training.
We investigate the fluctuations of the correlation dimension $D_2$ corresponding to the right eigenvectors of the layerwise Jacobians $\mathbf{W}^{l+1} \mathbf{D}^l$ based on Eq.\ (\ref{eq:frac_dim}) for various epochs during the training of \emph{FC10}.\footnote{For more epochs, please see the Supplementary Material.}
Our theoretical framework shows that the left eigenvectors of layerwise Jacobians of heavy-tailed networks are spatially multifractal in comparison to Gaussian random networks for which $\alpha=2$ and the eigenvalues of the layerwise Jacobians are delocalised.
Moreover, we find via simulations that training preserves the multifractal property of layerwise Jacobians when heavy-tailed initialisation is applied (Fig.~\ref{fig:dq_jac_transition_l=4_epoch=650}).
Although a perfect representation of delocalisation, i.e. $\langle D_q \rangle = 1$ for the Gaussian case (Fig.~\ref{fig:dq_jac_transition_l=4_epoch=650}(a) thick orange line) is set back by finite size effects and can only be achieved when the system size approaches infinity, the difference between delocalisation ($\alpha = 1.2$) and multifractality ($\alpha = 2.0$) is fully displayed; the fractal dimension $D_q$ corresponding to the heavy-tailed distribution, is a non-trivial function as it experiences a much more significant decrease with respect to $q$ relative to the Gaussian case.
Hence, it is sufficient to differentiate multifractality strength via $D_2$ which we have plotted on the $(\alpha, D_w^{1/\alpha})$ parameter space (Fig.~\ref{fig:dq_jac_transition_l=4_epoch=650}(b)); in particular, the average correlation dimension decreases with $\alpha$. 
The multifractal region also mostly emerges above the ordered transition line $(\alpha,\overline{D_w})$ which is consistent with the theoretical derivations \cite{Wardak2022}.

\subsection{Balanced contraction and expansion of internal neural representations} \label{sec:balance}

\begin{figure*}
\centering
\begin{overpic}[width=\textwidth]{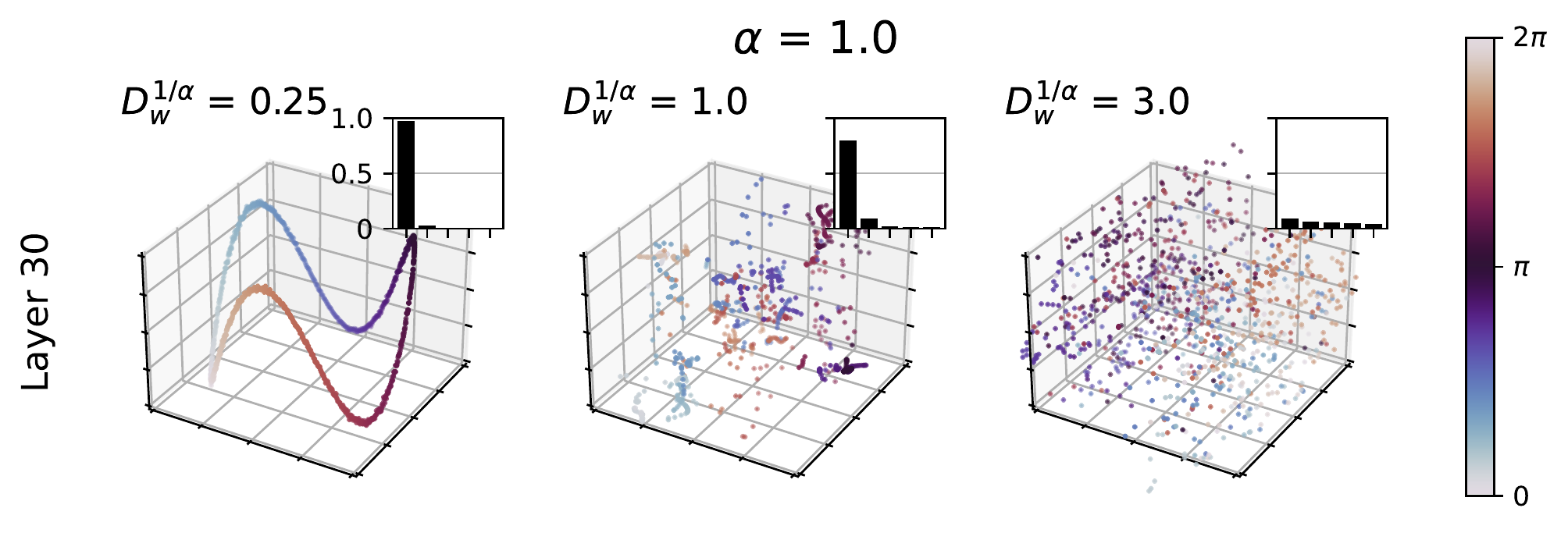}
\put (3.8,29.5) {\large\textbf{(a)}}
\put (32,29.5) {\large\textbf{(b)}}
\put (60.2,29.5) {\large\textbf{(c)}}
\end{overpic}
\caption{\label{fig:proj3d_single_alpha=100_cycli} \textbf{Propagation of manifold geometry through deep random networks.} Propagation of a great circle through FC20 initialised with $\alpha = 1.0$, projected to its three principal components (normalised); all 3 axes have a cut off $(-0.05,0.05)$.
Insets show the fraction of variance explained by the top 5 
singular values
with respect to the manifold represented at the corresponding layers in the subtitles.
The total variance and top singular value from left to right are respectively $(1.42 \times 10^{-10}, 1.38 \times 10^{-10}), (2.25 \times 10^4, 1.89 \times 10^4)$ \& $(7.70 \times 10^4, 6.79 \times 10^3)$.
The cyclic colourbar corresponds to the rotation angle $\theta$ of the input manifold; the colourbar is set between $[0, 2\pi]$.
}
\end{figure*}

We next analyse the computational advantages of the extended critical phase by considering the propagation of manifold geometry through deep random neural networks.
It has been shown \cite{Schoenholz2016} that random Gaussian feedforward networks may be trained precisely when information can propagate through them, which occurs at the classical edge of chaos \cite{lee2018deep, Poole2016}.
By applying Riemannian geometry to a random 1-dimensional circular manifold 
$\mathbf{h}^l(\theta)$ (with unit radius)
propagated through the layers, the classical ordered and chaotic regimes can be shown to correspond with the uniform compression and nonlinear expansion of internal neural representations respectively \cite{Poole2016}.
We study how this circular manifold propagates through random heavy-tailed DNNs in Fig.\ \ref{fig:proj3d_single_alpha=100_cycli}, finding that the ordered 
(Fig.\ \ref{fig:proj3d_single_alpha=100_cycli}(a))
and chaotic
(Fig.\ \ref{fig:proj3d_single_alpha=100_cycli}(c))
phases correspond to regimes of contraction and nonlinear expansion which respectively occur uniformly throughout the manifold.
Meanwhile, a combination of contraction and nonlinear expansion of input points is displayed along different parts of the manifold in the critical regime
(Fig.\ \ref{fig:proj3d_single_alpha=100_cycli}(b)).
Through principal component analysis (PCA), the principal components (PCs) show the proportion of signal which has been preserved (Fig.~\ref{fig:proj3d_single_alpha=100_cycli} insets).
In the chaotic regime, a majority of the signal, represented by the top 2 fraction of variance produces similar strengths as the noisy lower PCs.
On the other hand, the silent regime loses the signal due to contraction rather than noisy PC dispersion (top singular value is small, Fig.~\ref{fig:proj3d_single_alpha=100_cycli} caption).
Only the critical regime preserves the signal by avoiding contraction while maintaining a significant proportion of data in the first PC.%
\footnote{For more simulations on the circular manifold propagation with different values of $\alpha$, please see the Supplementary Material.}

\begin{figure*}
    \centering
    \includegraphics[width=\textwidth]{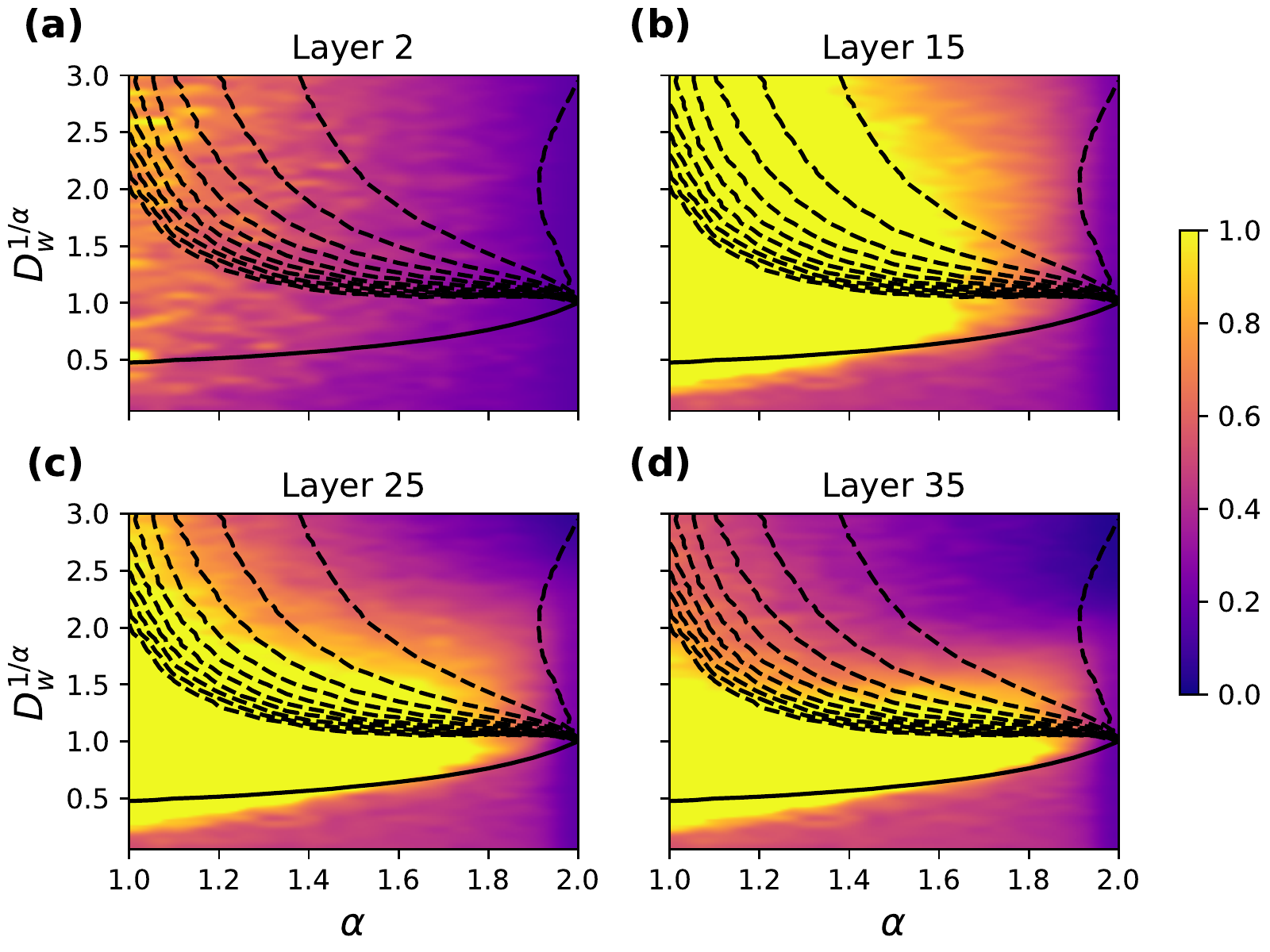}
    \caption{\textbf{The extended critical regime balances contraction and expansion.} A phase transition plot with the colormap representing the coefficient of variation of pairwise distances over an initially circular manifold propagated through a deep random network with 1000 neurons, averaged over 50 network ensembles.
    Overlaid are the theoretically predicted critical transition lines from Fig.\ \ref{fig:phase_transition}.
    The classical edge-of-chaos point appears at $(\alpha,D_w^{1/\alpha})=(2,1)$.
    }
    \label{fig:ge_mean_transition}
\end{figure*}

To compute the fluctuations of pairwise distances we use the dimensionless coefficient of variation
\begin{equation}
\mathrm{CV} = 
\frac{
\mathrm{std}_\theta(\Delta h^l(\theta))
}{
\langle\Delta h^l(\theta)\rangle_\theta
}
\end{equation}
in Fig.\ \ref{fig:ge_mean_transition},
where $\Delta h^l(\theta) := \norm{ \mathbf{h}^l(\theta + \Delta \theta) - \mathbf{h}^l(\theta)}$ is the change in Euclidean distance of the $l^{\text{th}}$ hidden layer as its input angle is perturbed by a small $\Delta \theta$; the average and standard deviation are taken over the angles $\theta \in [0, 2\pi)$.
We find that both the classical ordered and chaotic regimes have low fluctuations of pairwise distances, as each regime contracts and nonlinearly expands neural inputs uniformly across the ambient space and thus the manifold.
However, the extended critical regime displays large fluctuations of pairwise distances across the manifold and thus input space, demonstrating that the network qualitatively balances the contraction and expansion of neural inputs across space.
Such balanced contraction and expansion does not exist beforehand but emerges upon propagation by a number of layers in the network, as shown in Fig.\ \ref{fig:ge_mean_transition}(a).
As the circular manifold propagates deeper through the network, the region exhibiting simulataneous contraction and expansion evolves in a manner roughly following the analytically predicted continuous phase transition parameterised by $L$ (Fig.\ \ref{fig:ge_mean_transition}(b--d), black lines).
Particularly, the extended critical regime is consistent with that of mixed contraction and expansion.
The symmetry breaking caused by the simultaneous balancing of contraction and expansion throughout the neural manifold enables the propagation of information through many layers when networks are trained in this critical regime, extending the findings in \cite{Schoenholz2016} to heavy-tailed deep networks.
The observed fluctuations of the classical critical point ($(\alpha, D_w^{1/\alpha}) = (2,1)$) are smaller than deep in the extended critical phase (Fig.~\ref{fig:ge_mean_transition} (d)); this phenomenon might arise from a combination of finite-size effects and the localisations of the corresponding eigenvectors in each regime.

We hypothesise that the presence of multifractality in layerwise Jacobian eigenvectors allows for the balance of contraction and expansion to be enacted earlier in the layers of the deep network.
In the extended critical phase, the eigenvalues of a given eigenstate corresponding to successive layers are larger in modulus than those appearing at the silent transition line, quantified by the high Jacobian average.
This causes a subset of spatially multifractal directions in neural space to consistently appear above the critical line, allowing neural representations to experience expansion in those directions while contracting other directions, a form of symmetry breaking.
This balance between contraction in some directions and expansion in others allows neural networks to prioritise different parts of the input
when training on the problem data.
On the other hand, deep networks with Gaussian statistics have spatially delocalised layerwise Jacobians and gradients, so that each direction appears equivalent to the network even after quenching regardless of the problem data.
Consequently, Gaussian deep networks can only contract or nonlinearly expand and fold neural representations in all directions simultaneously without precision, corresponding to the classical ordered and chaotic regimes respectively \cite{Sompolinsky1988}.


\subsection{Heavy-tailed initialisation as a training strategy}

\begin{figure*}
\centering
\includegraphics[width=17.8cm,height=7.4cm]{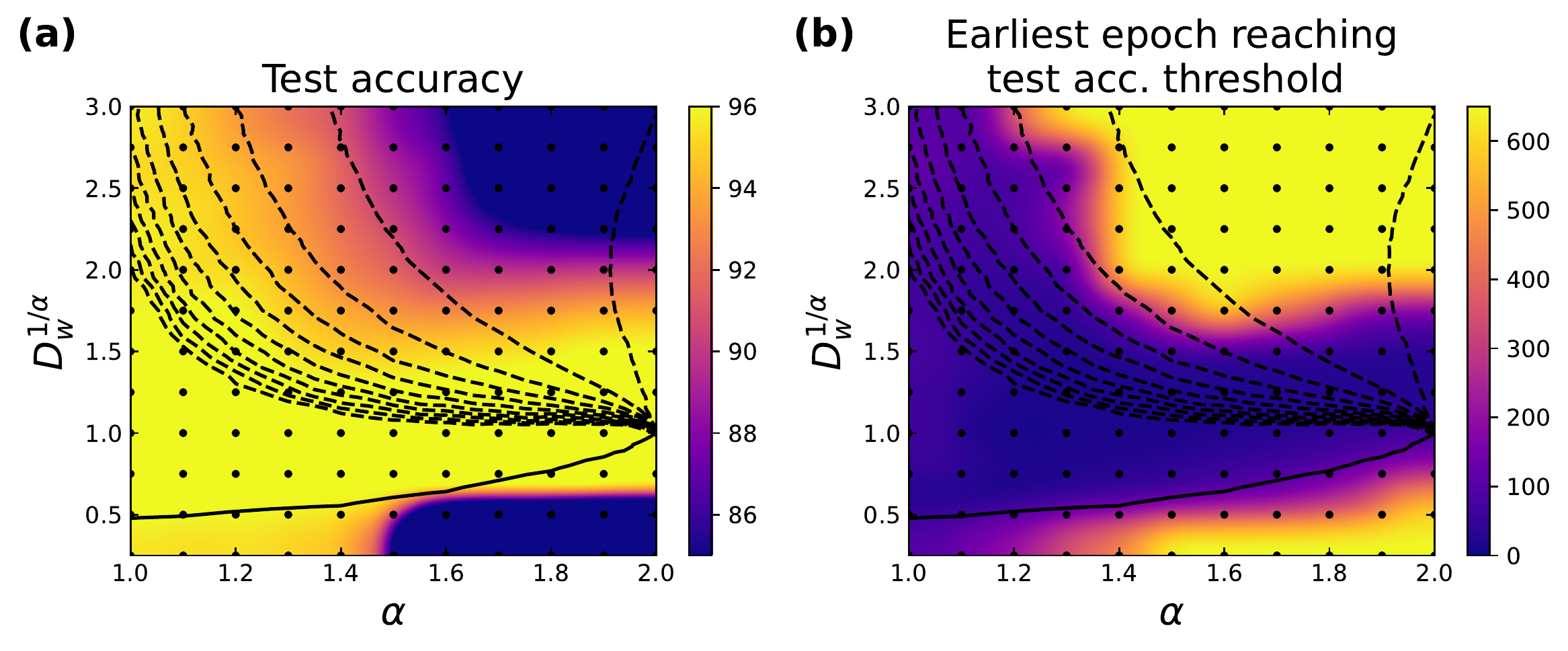}
\caption{\label{fig:fc10_grid_acc_and_ratio}
\textbf{The extended critical regime displays superior trainability and generalisation.}
(a) A colormap scatter plot displaying the testing accuracy of FC10 at epoch 650 with quadric interpolation.
The dots represent each realization at the initialisation parameters of the corresponding trained network.
Every network was trained for 650 epochs with a batch size of 1024 and learning rate of 0.001 with the vanilla SGD algorithm. 
(b) The same as in (a) but with the colormap representing the earliest epoch reaching the testing accuracy threshold of 93\% with a cutoff at epoch 650.
}
\end{figure*}

The relationship between the signal propagation of the network and its dynamic critical regime
suggests that deep networks initialised within the extended critical phase are easier to train.
To test this, we select parameters uniformly on a $(\alpha,D_w^{1/\alpha})$-grid and train \emph{FC10} for 650 epochs using standard SGD with learning rate 0.001 and batch size 128 (Fig.\ \ref{fig:fc10_grid_acc_and_ratio}).
Deep fully connected feedforward networks are known to suffer from the exploding and vanishing gradient problem under Gaussian initialisations far from the edge of chaos \cite{Schoenholz2016, 279181}.
We show in Fig.\ \ref{fig:fc10_grid_acc_and_ratio}(a) that such network architectures can be successfully trained when initialised in the extended critical regime, displaying a superior testing accuracy after 650 epochs.
The interval of successful training for Gaussian initialisations with $\alpha=2$ spans between $0.7\leq D_w^{1/\alpha} \leq1.5$ due to finite-size effects of the 784-neuron layers, with the interval closing into the classical edge-of-chaos point at $(\alpha, D_w^{1/\alpha})=(2,1)$ in the large network limit.
Computing the earliest epoch at which the testing accuracy threshold of 93\% is attained in Fig.\ \ref{fig:fc10_grid_acc_and_ratio}(b),
we find that the network converges to the successfully trained weight configuration significantly faster when initialised in the extended critical regime.
The change in training time due to differing initialisation parameters spans multiple orders of magnitude: networks initialised deep in the critical regime can be successfully trained in under 50 epochs, while deeply chaotic initialisation prevents the network from reaching a testing threshold of 93\% even after training is stopped at 650 epochs.

When initialised in the extended critical regime, deep neural networks train faster and attain higher testing accuracies than the chaotic and ordered regimes.
This vouches for the practical utility of the theoretical application of
heavy-tailed statistical physics
in the selection of parameters for the deep learning practitioner in real-world machine learning problems.
Since the regime is extended in parameter phase space, it is no longer necessary for the practitioner to fine-tune parameters to be specific values depending on the architecture or problem data.
From this perspective, our results provide a crucial guideline for the successful training and generalisation of deep neural networks regardless of problem domain.

\section{Discussion}

In this study, 
we have developed a novel mean-field theory for random deep neural networks with heavy-tailed weights, and used it to identify an extended critical regime of enhanced signal propagation which does not require extensive fine-tuning of parameters.
This extended critical regime with multifractal properties provides key computational advantages linked to balancing compression and expansion of internal neural representations, and therefore presents a theoretical framework for guiding parameter selection as well as design of deep networks in practice.

As shown in our study, the properties of the extended critical phase generalises the current analysis of singular values \cite{Martin2021} to a complete spectral theory of deep networks encompassing eigenvectors, and thus the spatial properties of the system dynamics, as well as eigenvalues \cite{Wardak2022}.
We have obtained this more general formulation through the use of dynamical systems theory, bypassing the singular values and opening up analysis on the spatially local properties of signal propagation via the eigenvectors of $\mathbf{D} \mathbf{W}$.
Therefore, we have established connections between the improved performance of DNNs, spectral properties of layerwise Jacobian matrices and self-organised criticality.
Our results are also universal in the sense that the precise distributions of weights are not required to be known apart from their tail asymptotics.
Additionally, the derived mean-field theory in Eq.\ (\ref{eq: iter_map}) applies to all sigmoidal activation functions, or more generally functions that are $o(\vert x \vert)$.

Our results on heavy-tailed DNNs indicate a change of view from the critical point or line requiring fine-tuning of parameters, towards a novel, extended critical regime. 
Previous studies have demonstrated that criticality between the ordered and chaotic phases enables effective information propagation across layers \cite{Schoenholz2016, Poole2016}, and prevents gradient vanishing and explosion from back-propagation, thus facilitating training. 
Aside from this functional advantage entailed by the edge of chaos, our work has demonstrated that extended criticality enables the balance of contraction and expansion of internal neural representations.
With the presence of multifractality which presents a mixture of localisation and delocalisation of eigenvectors, the trainability as well as generalisability of DNNs can be significantly improved as key features of the input can be better extracted. 
Since multifractality does not appear in the classical Gaussian case, precise extraction of local input features from internal neural space requires a precarious combination of delocalised eigenvectors for Gaussian DNNs but only a small robust sum of multifractal components in heavy-tailed DNNs.
These functional advantages of extended criticality, which does not require precise tuning of parameters, consequently illustrate a robust mechanism by which DNNs can possess remarkable performance in solving real-world problems. 
Our theory supplies the extended critical regime as a key guide for performing heavy-tailed initialization, thus providing a principled explanation of the empirical observation that DNNs with heavy-tailed singular spectra of weight matrices generalise better \cite{Martin2021}. 
As heavy-tailed weights emerge during the training process from L\'evy distributed gradients and gradient noise \cite{HodgkinsonM21, Chen2022}, our theory also provides a framework for understanding complex learning dynamics \cite{Chen2022, Feng2021}, which may result in better training algorithms for different learning tasks.

Finally, criticality underlies a wide range of biological systems ranging from families of proteins, networks of neurons to flocks of birds with crucial optimal computational capabilities and large dynamical repertoires \cite{Mora2011}. 
By linking heavy-tailed statistical physics with machine learning via random matrix theory, our new formalism on the concept of extended criticality may have general applicability to understanding these systems, suggesting that extended criticality with complex dynamics might be a general governing principle of biological and artificial intelligence.

\matmethods{
\subsection*{Code}
The code for generating all the simulations and figures can be found on the Github repository: \href{https://github.com/CKQu1/extended-criticality-dnn}{https://github/CKQu1/extended-criticality-dnn}.
\subsection*{Pretrained Networks}
The stable distribution fitting in Section 1A are performed on a total of 10 pretrained networks from torchvision.models of the Pytorch library (version 1.6.0), the network architects include: AlexNet, ResNet-18, ResNet-34, ResNet-50, ResNet-101, ResNet-152, ResNet101-32x4d, ResNet101-32x8d, Wide ResNet-50-2, Wide ResNet-101-2.
\subsection*{Other Networks}
The simulations conducted in the main text after Section A1 contain fully connected networks of various depths which all possess square connectivity matrices of size $784 \times 784$ (except for the final layer which is $784 \times 10$ due to the classification task on MNIST). 
We train the networks using vanilla SGD with all the hyperparameters specified above in the main text.
All the fully-connected networks were trained on the MNIST database.
}

\showmatmethods{} 

\acknow{The authors acknowledge the University of Sydney HPC service for providing high-performance computing that has contributed to the research results reported within this paper. This work was supported by the Australian Research Council (Grant Nos. DP160104316, DP160104368).
}

\showacknow{} 

\bibliography{pnas-sample}

\begin{thebibliography}{10}

\bibitem{Krizhevsky2012}
Krizhevsky A, Sutskever I, Hinton GE (2012) Imagenet classification with deep
  convolutional neural networks in {\em Advances in Neural Information
  Processing Systems}, eds.{} Pereira F, Burges CJC, Bottou L, Weinberger KQ.
\newblock (Curran Associates, Inc.), Vol.{}~25.

\bibitem{NEURIPS2020_1457c0d6}
Brown T, et~al. (2020) Language models are few-shot learners in {\em Advances
  in Neural Information Processing Systems}, eds.{} Larochelle H, Ranzato M,
  Hadsell R, Balcan MF, Lin H.
\newblock (Curran Associates, Inc.), Vol.{}~33, pp. 1877--1901.

\bibitem{yu2014automatic}
Yu D, Deng L (2014) {\em Automatic Speech Recognition - A Deep Learning
  Approach}.
\newblock (Springer).

\bibitem{Chialvo2010}
Chialvo DR (2010) Emergent complex neural dynamics.
\newblock {\em Nature Physics} 6(10):744--750.

\bibitem{LeCun2015}
LeCun Y, Bengio Y, Hinton G (2015) Deep learning.
\newblock {\em Nature} 521(7553):436--444.

\bibitem{Bahri2020}
Bahri Y, et~al. (2020) Statistical mechanics of deep learning.
\newblock {\em Annual Review of Condensed Matter Physics} 11(1):501--528.

\bibitem{Poole2016}
Poole B, Lahiri S, Raghu M, Sohl-Dickstein J, Ganguli S (2016) Exponential
  expressivity in deep neural networks through transient chaos in {\em
  Proceedings of the 30th International Conference on Neural Information
  Processing Systems}.
\newblock (Curran Associates Inc., Barcelona, Spain), p. 3368–3376.

\bibitem{Pennington2018}
Pennington J, Schoenholz S, Ganguli S (2018) The emergence of spectral
  universality in deep networks in {\em Proceedings of the Twenty-First
  International Conference on Artificial Intelligence and Statistics}, eds.{}
  Storkey A, Perez-Cruz F.
\newblock (PMLR, Proceedings of Machine Learning Research), Vol.{}~84, pp.
  1924--1932.

\bibitem{Schoenholz2016}
{Schoenholz} SS, {Gilmer} J, {Ganguli} S, {Sohl-Dickstein} J (2016) {Deep
  Information Propagation}.
\newblock {\em arXiv e-prints} p. arXiv:1611.01232.

\bibitem{Glorot2010}
Glorot X, Bengio Y (2010) Understanding the difficulty of training deep
  feedforward neural networks in {\em Proceedings of the Thirteenth
  International Conference on Artificial Intelligence and Statistics},
  Proceedings of Machine Learning Research, eds.{} Teh YW, Titterington M.
\newblock (PMLR, Chia Laguna Resort, Sardinia, Italy), Vol.{}~9, pp. 249--256.

\bibitem{simsekli19a}
Simsekli U, Sagun L, Gurbuzbalaban M (2019) A tail-index analysis of stochastic
  gradient noise in deep neural networks in {\em Proceedings of the 36th
  International Conference on Machine Learning}, Proceedings of Machine
  Learning Research, eds.{} Chaudhuri K, Salakhutdinov R.
\newblock (PMLR), Vol.{}~97, pp. 5827--5837.

\bibitem{HodgkinsonM21}
Hodgkinson L, Mahoney MW (2021) Multiplicative noise and heavy tails in
  stochastic optimization in {\em Proceedings of the 38th International
  Conference on Machine Learning, {ICML} 2021, 18-24 July 2021, Virtual Event},
  Proceedings of Machine Learning Research, eds.{} Meila M, Zhang T.
\newblock ({PMLR}), Vol.{} 139, pp. 4262--4274.

\bibitem{Martin2021}
Martin CH, Peng TS, Mahoney MW (2021) Predicting trends in the quality of
  state-of-the-art neural networks without access to training or testing data.
\newblock {\em Nature Communications} 12(1):4122.

\bibitem{Nolan2020}
Nolan JP (2020) {\em {Basic Properties of Univariate Stable Distributions}}.
\newblock (Springer International Publishing), pp. 1--23.

\bibitem{Samorodnitsky_Taqqu}
Samorodnitsky G (1994) {\em Stable non-Gaussian random processes : stochastic
  models with infinite variance}.
\newblock (New York : Chapman \& Hall, [1994] ©1994).

\bibitem{Bordenave2011}
Bordenave C, Caputo P, Chafaï D (2011) Spectrum of non-hermitian heavy tailed
  random matrices.
\newblock {\em Communications in Mathematical Physics} 307(2):513.

\bibitem{Wardak2021}
Wardak A, Gong P (2021) Fractional diffusion theory of balanced heterogeneous
  neural networks.
\newblock {\em Phys. Rev. Research} 3(1):013083.

\bibitem{lee2018deep}
Lee J, et~al. (2018) Deep neural networks as gaussian processes in {\em
  International Conference on Learning Representations}.

\bibitem{Rajan2006}
Rajan K, Abbott LF (2006) Eigenvalue spectra of random matrices for neural
  networks.
\newblock {\em Phys. Rev. Lett.} 97(18):188104.

\bibitem{Aljadeff2015}
Aljadeff J, Stern M, Sharpee T (2015) Transition to chaos in random networks
  with cell-type-specific connectivity.
\newblock {\em Phys. Rev. Lett.} 114(8):088101.

\bibitem{Evers2008}
Evers F, Mirlin AD (2008) Anderson transitions.
\newblock {\em Rev. Mod. Phys.} 80(4):1355--1417.

\bibitem{Wardak2022}
{Wardak} A, {Gong} P (2022) {Extended Anderson Criticality in Heavy-Tailed
  Neural Networks}.
\newblock {\em arXiv e-prints} p. arXiv:2202.05527.

\bibitem{Bordenave2012}
Bordenave C, Chafaï D (2012) {Around the circular law}.
\newblock {\em Probability Surveys} 9(none):1 -- 89.

\bibitem{LucasMetz2019}
Lucas~Metz F, Neri I, Rogers T (2019) Spectral theory of sparse non-hermitian
  random matrices.
\newblock {\em Journal of Physics A: Mathematical and Theoretical}
  52(43):434003.

\bibitem{Farrell2019}
Farrell M, Recanatesi S, Moore T, Lajoie G, Shea-Brown E (2019) Recurrent
  neural networks learn robust representations by dynamically balancing
  compression and expansion.
\newblock {\em bioRxiv} p. 564476.

\bibitem{Moretti2013}
Moretti P, Muñoz MA (2013) Griffiths phases and the stretching of criticality
  in brain networks.
\newblock {\em Nature Communications} 4(1):2521.

\bibitem{Sompolinsky1988}
Sompolinsky H, Crisanti A, Sommers HJ (1988) Chaos in random neural networks.
\newblock {\em Phys. Rev. Lett.} 61(3):259--262.

\bibitem{279181}
Bengio Y, Simard P, Frasconi P (1994) Learning long-term dependencies with
  gradient descent is difficult.
\newblock {\em IEEE Transactions on Neural Networks} 5(2):157--166.

\bibitem{Chen2022}
Chen G, Qu CK, Gong P (2022) Anomalous diffusion dynamics of learning in deep
  neural networks.
\newblock {\em Neural Networks} 149:18--28.

\bibitem{Feng2021}
Feng Y, Tu Y (2021) The inverse variance-flatness relation in stochastic
  gradient descent is critical for finding flat minima.
\newblock {\em Proceedings of the National Academy of Sciences}
  118(9):e2015617118.

\bibitem{Mora2011}
Mora T, Bialek W (2011) Are biological systems poised at criticality?
\newblock {\em Journal of Statistical Physics} 144(2):268--302.

\end{thebibliography}

\end{document}


\nolinenumbers


\maketitle




\begin{figure*}
\centering
\begin{overpic}[width=\textwidth]{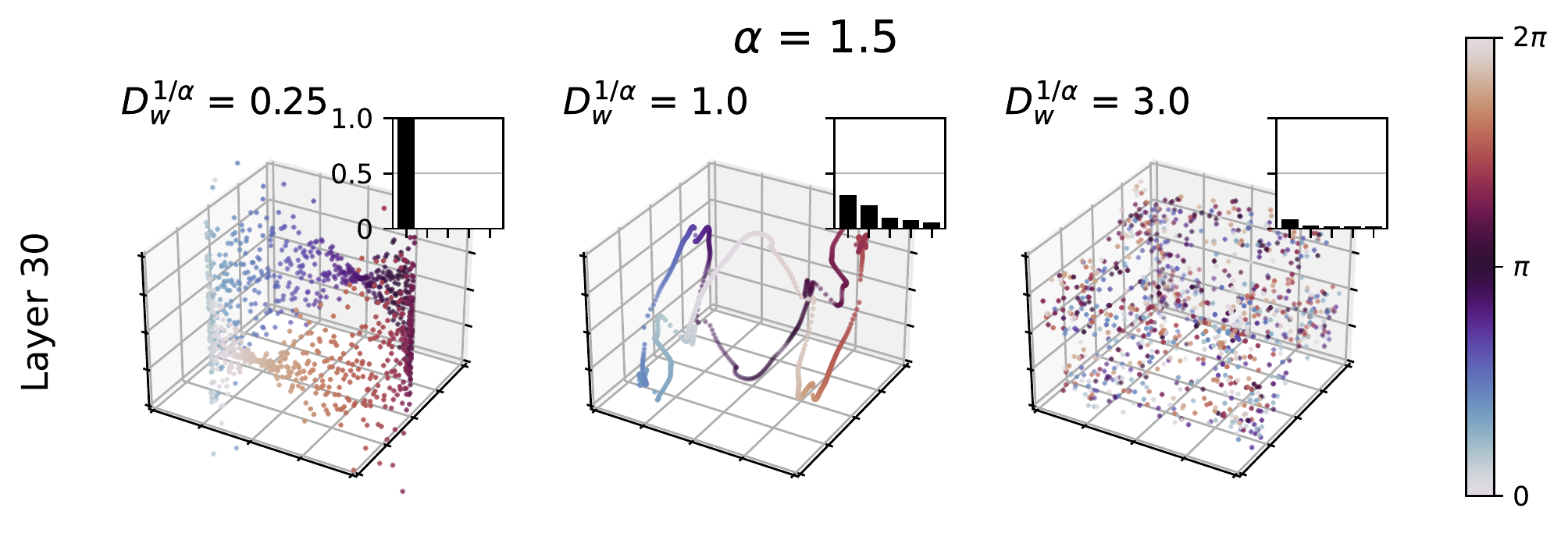}
\put (0,34) {\large\textbf{(a)}}
\end{overpic}
\begin{overpic}[width=\textwidth]{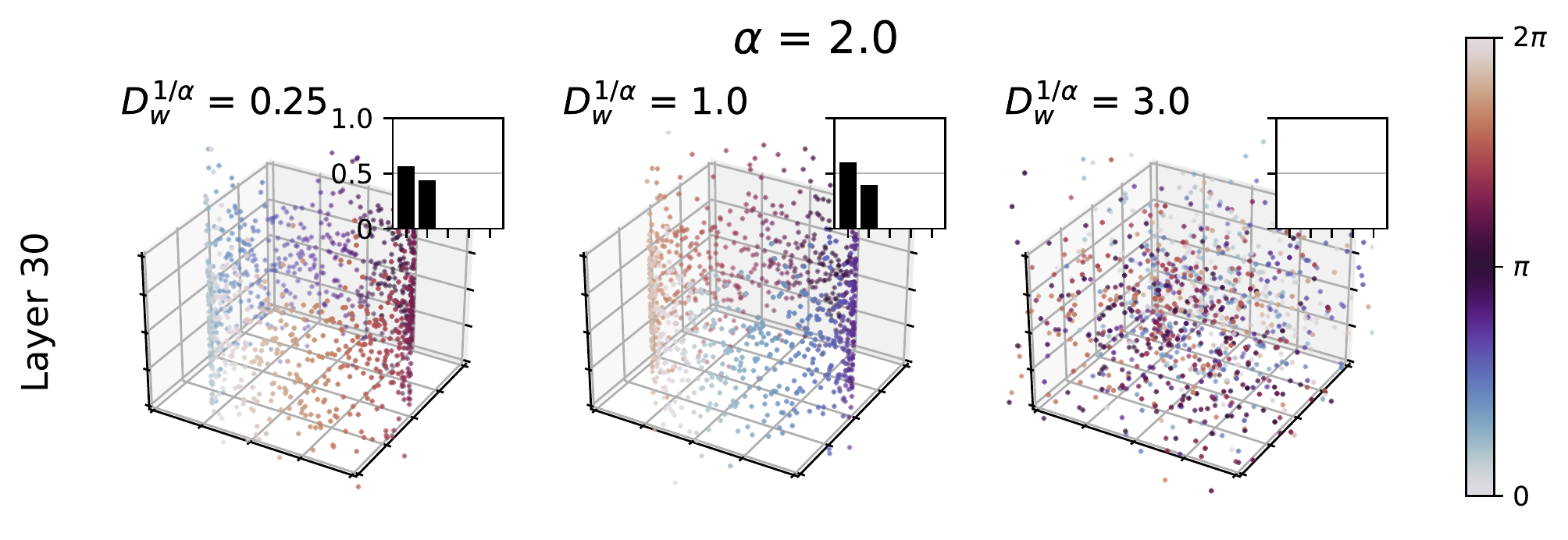}
\put (0,34) {\large\textbf{(b)}}
\end{overpic}
\caption{\label{fig:proj3d_single_alpha=100_cycli} \textbf{Propagation of manifold geometry through deep random networks.} 
(a) Propagation of a great circle through FC20 initialized with $\alpha = 1.5$, projected to its three principal components (normalised); all 3 axes have a cut off $(-0.05,0.05)$.
Insets show the fraction of variance explained by the top 5 singular values with respect to the manifold represented at the corresponding layers in the subtitles.
The total variance and top singular value from left to right are respectively $(7.70 \times 10^{-18}, 7.68 \times 10^{-18}), (1.46 \times 10^3, 4.39 \times 10^2)$ \& $(5.17 \times 10^4, 4.25 \times 10^2)$.
The cyclic colorbar corresponds to the rotation angle $\theta$ of the input manifold; the colorbar is set between $[0, 2\pi]$.
(b) Same as in (a) but for $\alpha = 2.0$.
The total variance and top singular value from left to right are respectively $(7.41 \times 10^{-24}, 4.18 \times 10^{-24}), (1.24 \times 10^{-5}, 7.47 \times 10^{-6})$ \& $(4.28 \times 10^4, 3.10 \times 10^2)$.
}
\end{figure*}

\begin{figure}
\centering
\begin{tabular}[b]{cc}
\begin{tabular}[b]{c}
\begin{overpic}[width=0.45\textwidth]{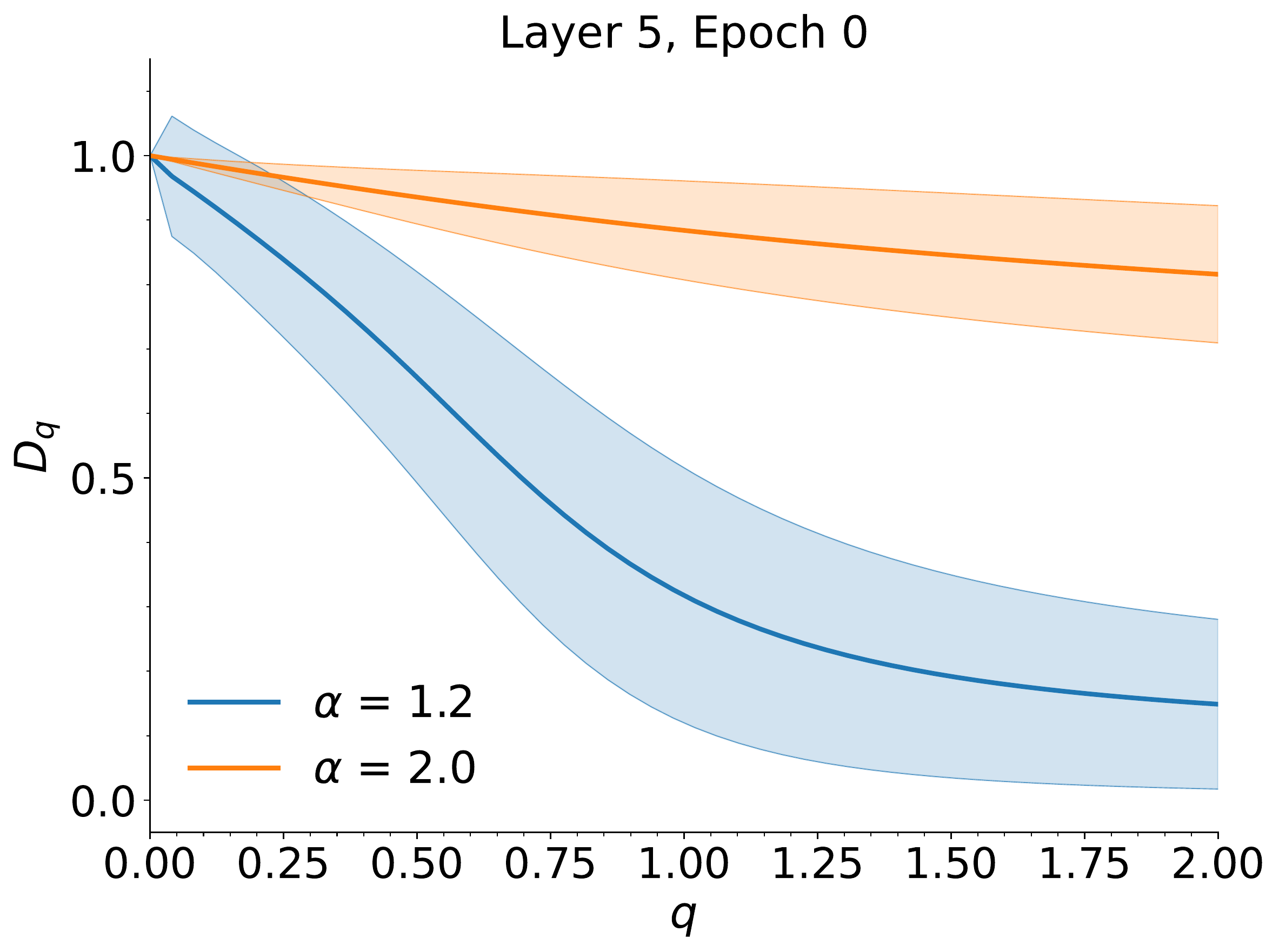}
\put (0,72) {\large\textbf{(a)}}
\end{overpic} \\
\begin{overpic}[width=0.45\textwidth]{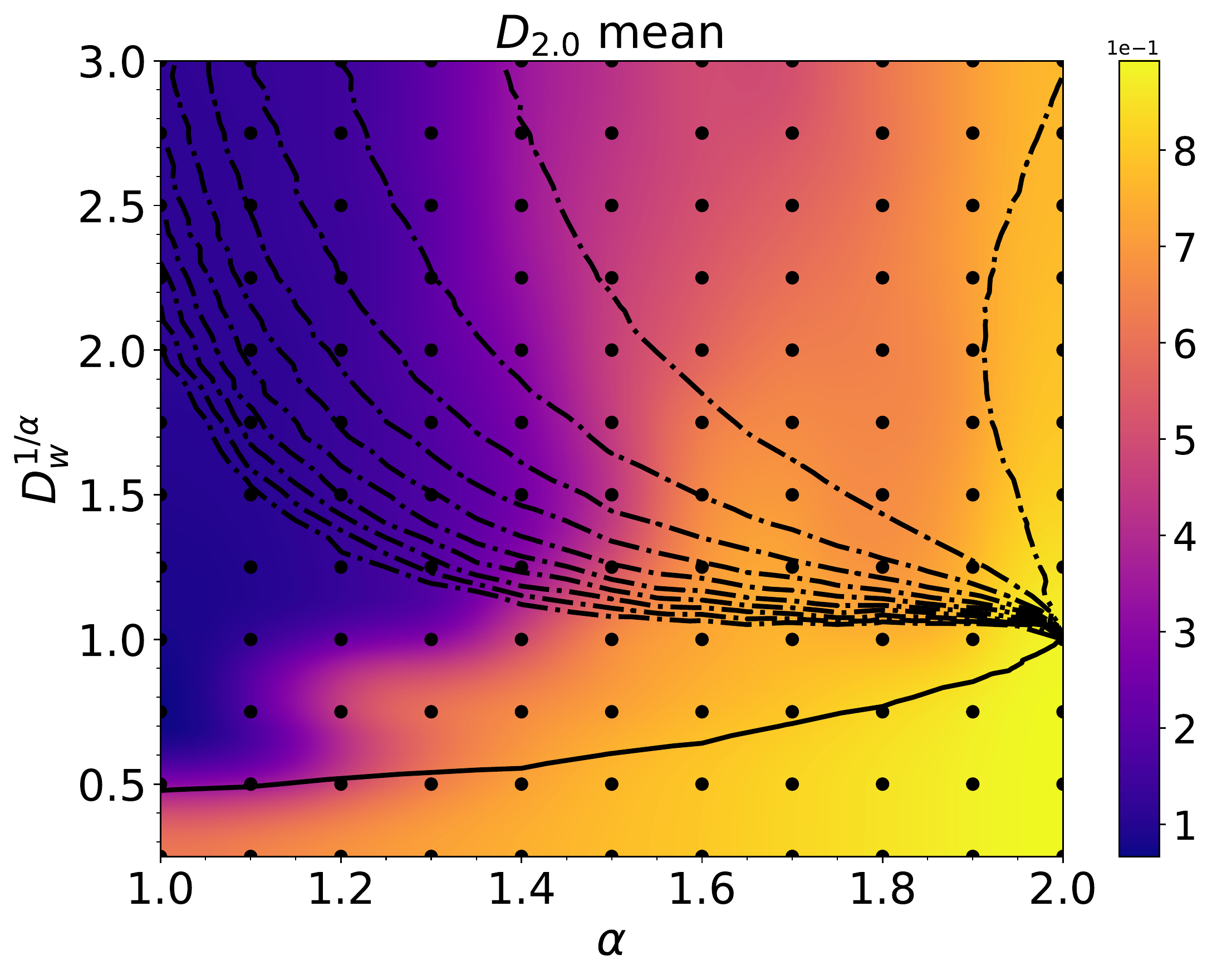}
\put (0,79) {\large\textbf{(c)}}
\end{overpic}
\end{tabular}
\begin{tabular}[b]{c}
\begin{overpic}[width=0.45\textwidth]{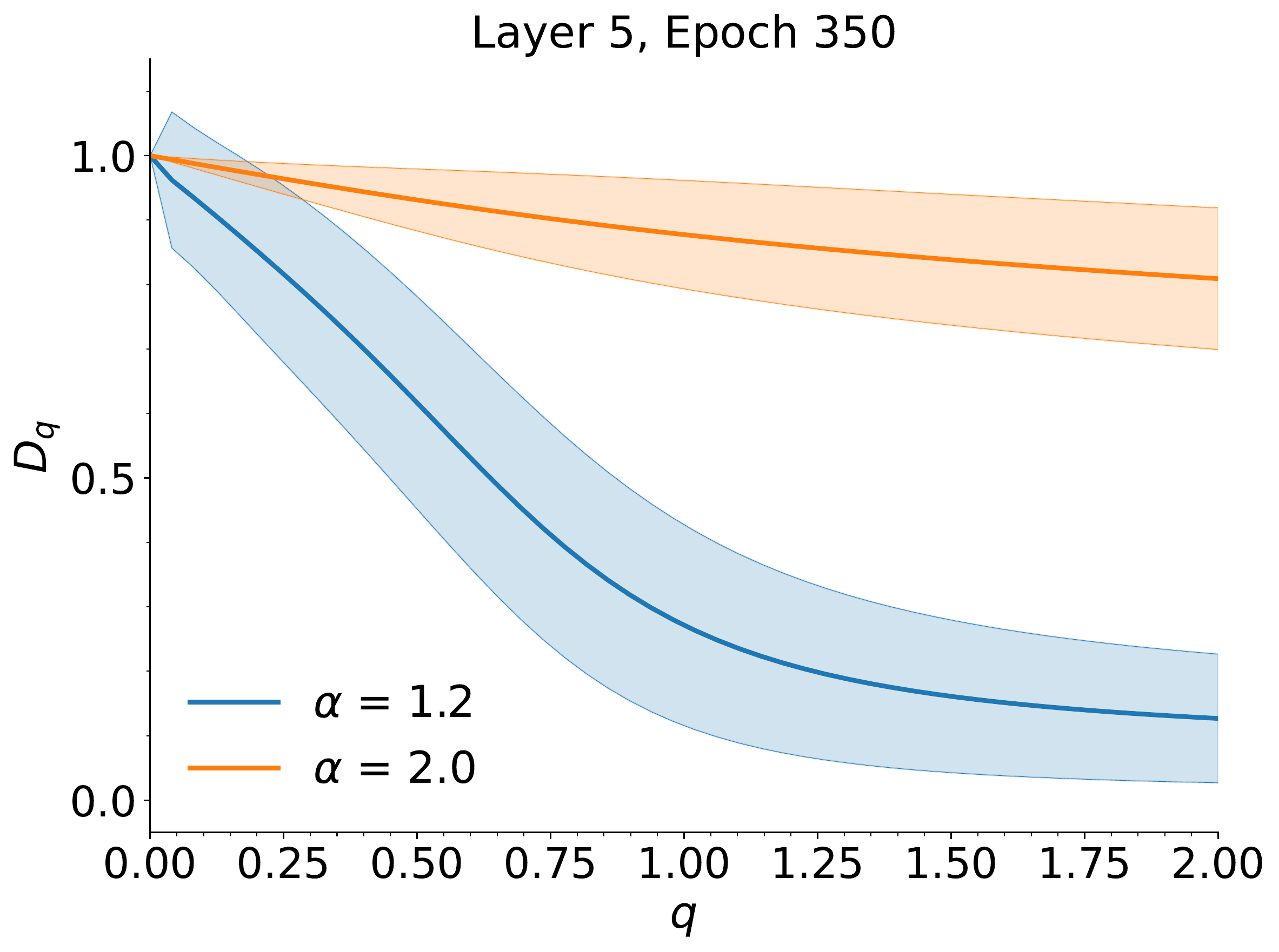}
\put (0,72) {\large\textbf{(b)}}
\end{overpic} \\
\begin{overpic}[width=0.45\textwidth]{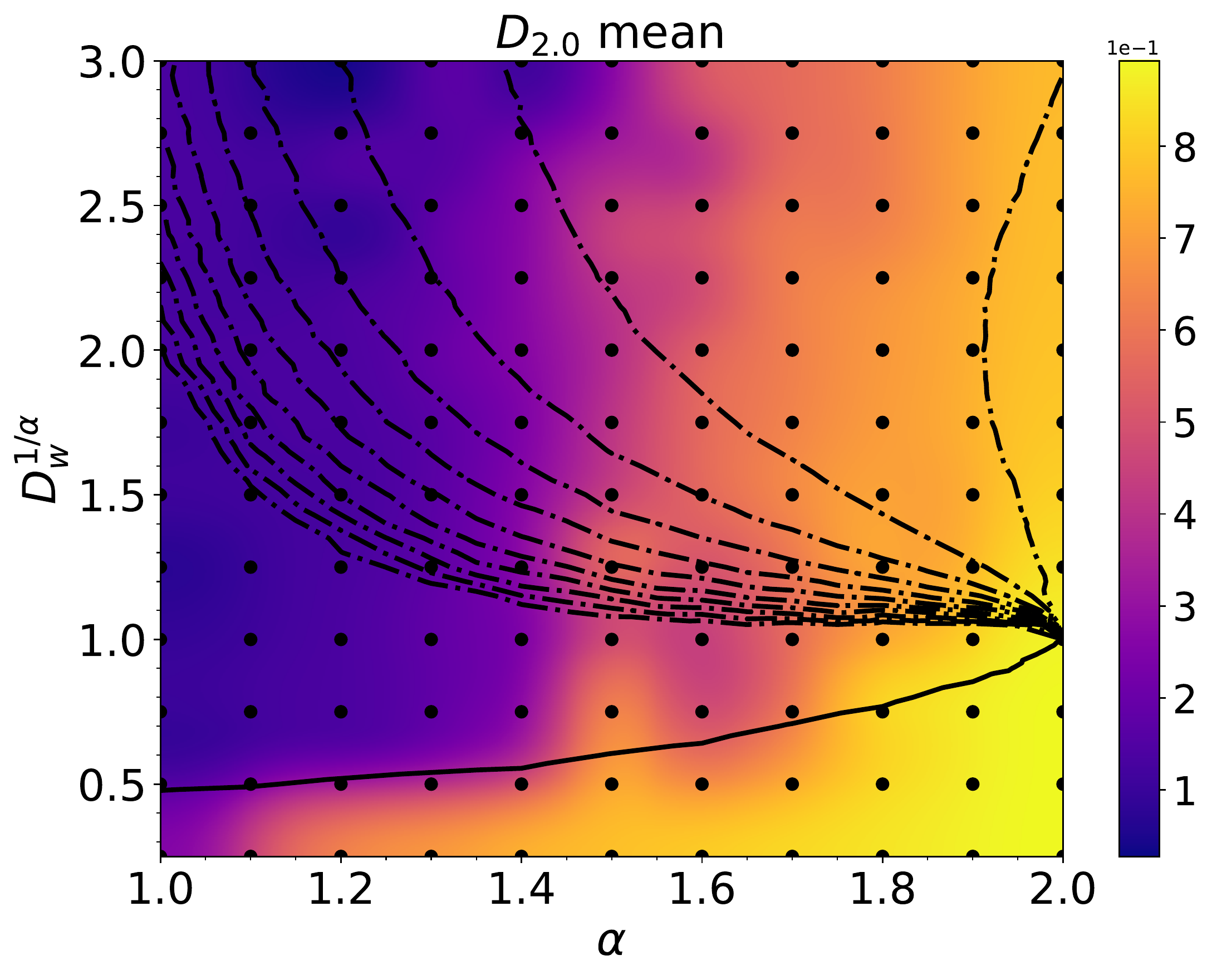}
\put (0,79) {\large\textbf{(d)}}
\end{overpic}
\end{tabular}
\end{tabular}
\caption{\label{fig:dq_jac_transition_l=4_epoch=650} \textbf{Fractal dimension $D_q$ vs $q$}. 
(a) The mean fractal dimension $D_q$ of the right eigenvectors of $\mathbf{W}^6 \mathbf{D}^5$ plotted with its standard deviation respectively for $\alpha = 1.2$ and $\alpha = 2.0$ with $D_w^{1/\alpha} = 1.5$ at epoch 0 (before training) in both cases.
(b) Same as in (a) but for epoch 350.
(c) The mean fractal dimension across all right eigenvectors of $\mathbf{W}^6 \mathbf{D}^5$ at epoch 650 plotted on the phase overlaid with the phase transition diagram in the main text (Fig.\ 2)
(d) Same as in (c) but for epoch 350.
}
\end{figure}

\FloatBarrier
